\documentclass{article}
\usepackage{ijcai22}

\usepackage{times}
\usepackage{soul}
\usepackage{url}
\usepackage[hidelinks]{hyperref}
\usepackage[utf8]{inputenc}
\usepackage[small]{caption}
\usepackage{graphicx}
\usepackage{amsmath}
\usepackage{amsthm}
\usepackage{booktabs}
\usepackage{algorithm}
\usepackage{algorithmic}
\title{Can counterfactual explanations of AI systems' predictions skew lay users' causal intuitions about the world? If so, can we correct for that?}

\author{
    Marko Te\v{s}i\'{c}
    \and
    Ulrike Hahn
    \affiliations
    Birkbeck, University of London
    \emails
    \{m.tesic, u.hahn\}@bbk.ac.uk
}

\begin{document}

\maketitle

\begin{abstract}
  Counterfactual (CF) explanations have been employed as one of the modes of explainability in explainable AI---both to increase the transparency of AI systems and to provide recourse. Cognitive science and psychology have pointed out that people regularly use CFs to express causal relationships. Most AI systems, however, are only able to capture associations or correlations in data so interpreting them as casual would not be justified. In this paper, we present two experiments (total $N=364$) exploring the effects of CF explanations of AI system's predictions on lay people's causal beliefs about the real world. In Experiment 1 we found that providing CF explanations of an AI system's predictions does indeed (unjustifiably) affect people's causal beliefs regarding factors/features the AI uses and that people are more likely to view them as causal factors in the real world. Inspired by the literature on misinformation and health warning messaging,  Experiment 2 tested whether we can correct for the unjustified change in causal beliefs. We found that pointing out that AI systems capture correlations and not necessarily causal relationships can attenuate the effects of CF explanations on people's causal beliefs.
\end{abstract}

\section{Introduction}

Interest in automatically-generated explanations for predictive AI systems has grown considerably in recent years \cite{DARPA,doshi2017towards,gunning2019darpa,montavon2018methods,rieger2018structuring,samek2017explainable}. It is argued that explanations provide transparency for what are often black-box procedures and transparency is viewed as critical for fostering the acceptance of AI systems in real-world practice \cite{bansal2014towards,chen2014situation,fallon2018improving,hayes2017improving,mercado2016intelligent,wachter2017counterfactual}. Explainable AI (XAI) has emerged as a field to address this need for AI systems' predictions to be followed by explanations of these predictions.

Common approaches to (post hoc) explainability of specific predictions of AI systems include feature importance \cite{lundberg2017unified,ribeiro2016should}, saliency maps \cite{simonyan2013deep}, and example-based methods \cite{koh2017understanding}. In this paper we focus on counterfactual (CF) explanations of specific predictions of AI systems \cite{forster2021capturing,galhotra2021explaining,karimi2020survey,kommiya2021towards,lucic2020does,poyiadzi2020face,van2021evaluating,verma2020counterfactual,wachter2017counterfactual}. These explanations describe changes in an AI system's inputs (features/factors) that alter the AI system's output (prediction/label) and lead to favorable outputs. CFs explanations address questions such as `Why A rather than B?'; for example, `Why did the AI system deny the loan rather than approve it?'. An answer to this questions would be a CF explanation:~`If Tom's salary had been at least £30k, the AI system would have offered him a loan'. CF explanations not only provide us with an insight into why an AI system made a certain prediction (`deny the loan'), but also what a user can do in order to flip the prediction (`offer a loan'). In other words, CFs explanation may also be able to provide recourse for users \cite{karimi2020survey}. Furthermore, CF explanations naturally embody contrastiveness, i.e.~the ability to address the question of why this prediction instead of some other one, which is one of the attributes that people expect explanations to have \cite{miller2019explanation}.

A significant body of research on CF explanations can be found in cognitive science and psychology. Some of the results of this research suggest that CF explanations often convey causal relations \cite{byrne2007precis,byrne2016counterfactual,byrne2019counterfactuals,thompson2002reasoning} and that making causal judgments often requires comparing actual and relevant counterfactual situations \cite{gerstenberg2021counterfactual,wells1989mental}. For example, taking painkillers can have side effects such as fatigue. In a situation where a runner sprained an ankle and took painkiller \emph{A} which has fatigue as one of its side effects, people would judge that painkiller \emph{A} has caused poor performance and loss of the race when they are aware of an alternative painkiller \emph{B} without side effects. Here people formed a CF:~if the runner had taken painkiller \emph{B}, she would not have had the side effects. However, when painkiller \emph{B} also leads to side effects, people judge painkiller \emph{A} to have less causal impact on the race outcome:~even if the runner had taken \emph{B}, she still would have had side effects \cite{mccloy2002semifactual}.

AI systems are typically predictive in nature and are capturing associations and correlations in data, not causal processes that generated the data. More specifically, in most applications of AI systems we use data \textbf{X} and $Y$ to estimate a function $f$, which in turn is used to generate predictions $\hat{Y}$ for new instances. No underlying theoretical causal model for function $f$ is assumed. Moreover, $f$ is not expected to adequately capture the underlying (causal) processes or real-world mechanisms that generated the data used for training and estimation. It is thus entirely possible that explanations for predictions $\hat{Y}$ that comprise of (changes) in features \textbf{X} have no clear causal connection (when, for example, \textbf{X} contains heavily engineered features) or have an anti-causal relationship, where $Y$ is a cause of some \textbf{X}. Furthermore, due to regularization it is possible that some of the actually causal \textbf{X} are left out or their impact on estimating $\hat{Y}$ is reduced \cite{del2021prediction}. One should then be careful when using AI systems and explanations of their predictions not to misinterpret AI systems in a causal manner and be wary of their limits \cite{medium_2021,molnar2020pitfalls}. This, however, may be easier said then done, particularly in the case of CF explanations of AI systems' predictions. CFs explanations normally assume that the change in feature values maps onto the actions in the real world. This implies that CFs explanation ought to incorporate a causal mechanism that would allow the person receiving the explanation to meaningfully intervene in the real world \cite{barocas2020hidden}. Some work on incorporating the real-world causal relationships in CF explanations has been done \cite{karimi2020model,karimi2020survey}, however the vast majority of CF explanation generation algorithms do not account for the causal structure of the world \cite{verma2020counterfactual}.

If people naturally associate CF with causal reasoning as is suggested by the psychological and cognitive science research, then they may be especially prone to slipping into causal interpretations of AI systems results when they are presented with CF explanations. As a consequence, they may form an (unjustified) mental model of the causal structure of the world or the underlying processes that generated the data. In other words, it may lead the recipients of CF explanations to form disingenuous and over attributive perspectives with respect to these systems. Recent empirical work suggests that CFs explanations do promote causal interpretations of features/factors used by an AI system \cite{warren2022features}, making plausible the claim that people may indeed form over attributive perspective regarding AI systems predictions when coupled with CF explanations. 

In this paper we test the possibility that CF explanations may lead lay people into believing that relations captured by AI systems are causal in the real world. We report two experiments. The first investigated whether lay people are more likely to form causal beliefs about the factors/features AI systems are using when these are presented with CF explanations. The second experiment, explores a possible means to prevent lay people from forming inadvertent causal beliefs due to CF explanations.

\section{Experiment 1}

The aim of this experiment was to explore lay people's causal beliefs after having received a prediction made by an AI system, which is then supplemented with a CF explanation. The main hypothesis is that people's causal beliefs about the world will be (unjustifiably) affected by CF explanations of AI system's predictions. More specifically, we hypothesise that people will erroneously hold beliefs that the features an AI has used to make predictions are more causal when a CF explanation of the AI system's prediction is provided, compared to when the prediction of an AI system is presented without a CF explanation, and compared to a baseline (where no AI system or its predictions are mentioned).

As AI systems are predictive in nature, one might argue that the above hypothesised effect may be due to lay people conflating the prediction/predictive power of AI systems with causation. The second hypothesis is aimed at testing this possibility. More specifically, we hypothesize that knowing an AI system is using certain feature A to predict label B and knowing what the predictions are \emph{will} change people's \emph{expectation} as to how good a predictor feature A is with respect B, compared to the baseline. Crucially, however, we hypothesize that additionally knowing an explanation for that prediction \emph{will not} further change people's \emph{expectation} as to how good a predictor A is. This finding would imply that any change in \emph{causal beliefs} would be due to the presence of a CF explanation of AI predictions and cannot be accounted for by a change in \emph{expectation} of how good the features the AI system uses are in predicting the label. Experiment 1 tested both hypotheses.

\subsection{Methods}

\paragraph{Participants}\label{s: part1}

A total of 93 participants ($N_{\mathrm{female}} = 74$, one participant identified as neither male nor female, $M_{\mathrm{age}} = 37.2$, $SD=13.3$) were recruited from Prolific Academic (\url{www.prolific.ac}). All participants were native English speakers residing in the UK or Ireland whose approval ratings were 95\% or higher. They gave informed consent and were paid £6.24 an hour rate for partaking in the study, which took on average 8.1 min to complete. Both Experiment 1 and 2 were approved by the the Department of Psychological Sciences, Birkbeck, University of London Ethics Committee (reference 2021074).

\paragraph{Design}
Participants were randomly assigned to one of three between-participant groups:~the Control/baseline group where participants were asked about their intuitions regarding how certain factors/features influence salary without mentioning AI systems or explanations of AI systems ($N = 30$); the AI Prediction group where participants were told about the AI system and the features it uses as well as what the prediction is ($N = 31$); and, the AI Explanation group where they were told about AI system, features, what prediction is, and received a CF explanation of the prediction for each feature ($N = 32$).

The experiment had three dependent variables:~Expectation, Confidence, and Action. Expectation dependent variable measured people's beliefs regarding how well features predict the label. Confidence dependent variable measured how confident participants were in their expectation estimates. The main reason for including Confidence dependent variable was to disentangle between people's beliefs about the predictive power of the features and their confidence in how predictive they believe the features are. We do not, however, have any hypotheses as to how Confidence will change as a function of the group people were assigned to. Lastly, Action dependent variable measured people's causal beliefs about the real world in terms of their willingness to act or recommend a certain action to be done in the real world. All participants provided answers for each dependent variable.

\paragraph{Materials}

To test the hypotheses we used salary as a domain; it is reasonable to expect that most  participants will have some familiarity regarding factors/features affecting salary and that they would already have developed certain intuitions about these factors. We chose 9 factors/features that are to various extents intuitively related to higher/lower salary. These were:~education level, the sector the employee works in (private or public), the number of hours of sleep, whether or not the employee owns a smart watch, whether or not the employee owns an office plant, whether or not the employee gets expensive haircuts, whether or not the employee wears expensive clothes, whether or not the employee goes skiing multiple times a year, and whether or not the employee rents a penthouse apartment. We aimed to have a range of factors/features whereby some are intuitively causing higher/lower salary (e.g.~education level, sector), some are intuitively not relevant to salary (e.g.~office plant, smart watch), and some are potential consequences or effects of higher salary rather than causing higher salary (e.g.~expensive clothes, expensive haircuts, renting penthouse apartments). With these factors/features we sought to cover possible ranges of Expectation and Action dependent variables. Namely, we hoped that for some factors/features such as education level both Expectation estimates and Action estimates would be high (i.e.~education level is a good predictor of salary and to increase their salary one might consider getting a higher degree); some factors/features would have both Expectation and Action estimates very low (e.g.~whether or not someone has an office plant does not seem to be related to salary and buying an office plant to increase salary would seem like a futile endeavour); lastly, some factors/features such as expensive clothes and renting a penthouse apartment would have higher Expectation estimates but lower Action estimates (that someone is renting a penthouse apartment may be an indicator that they have a high salary, but one would not presumably rent a penthouse apartment because they believe that would increase their salary). The features/factors were not chosen from a specific data set, but were devised for the purposes of the experiment.

All collected participant data and materials as well as the analysis code are available via OSF:~\url{https://osf.io/xu7v6/?view_only=a4d11733f3a546cca4b76ad8fbc75018}. 

\paragraph{Procedure}\label{s: procedure}

After  providing informed consent and  basic demographic information (age, gender, and first language; no personally identifiable information was collected), participants were shown a welcome message. Participants then answered two preliminary questions:~`How familiar are you with the factors that may affect salary?' and `How familiar are you with the AI technology, e.g.~AI systems that are able to make predictions?'. Answers to both questions were  on a 7-point Likert scale from `1 - Not at all familiar' to `7 - Extremely familiar'. The main motivation for including these questions was to check whether any differences among the three groups in the subsequent Expectation or Action estimates were due to differences in familiarity with the domain (salary) or familiarity with AI technology.

Following these two preliminary questions, participants saw a preamble for the specific group they were assigned to, i.e.~Control, AI Prediction, AI Explanation (square brackets indicate which text was presented to which group):

\vspace{1em}

\begin{itemize}
    \item[] Your good friend Tom is looking to increase his \textbf{salary}. He’s asked you for advice on how to best achieve that. [all three groups]
    \item[] There are a range of factors that are related to a higher salary. You will now consider some of these factors. [only the Control group]
    \item[] In your search for ways to help your friend you have found an \textbf{AI system} that can predict whether people’s yearly \textbf{salaries} are \textit{higher than/equal to £30k ($\geq$ £30k)} or \textit{lower than £30k ($<$ £30k)}. [AI Prediction and AI Explanation groups]
    \item[] The AI system uses a number of \textbf{factors} to make the prediction. You do not know, however, how much each factor is important for the AI system when it is making its predictions. [only the AI Prediction group]
    \item[] The AI system uses a number of \textbf{factors} to make the prediction. The AI system also has an option to provide you with \textbf{explanations} regarding its predictions. [only the AI Explanation group]
    \item[] [NEXT PAGE]
    \item[] You input Tom’s details for all factors into the AI system and it predicts that his yearly salary is \textit{\textbf{lower than £30k ($<$ £30k)}}. [AI Prediction and AI Explanation groups]
    \item[] The AI system now provides you with explanations with respect to each factor as to why it predicts that Tom’s salary is lower than £30k ($<$ £30k). [only the AI Explanation group]
\end{itemize}


The cutoff £30k was used as that figure was close to the median salary in the UK in 2020. After participants read the preamble for the group they were assigned to, they proceeded to answer the three questions (Expectation, Confidence, Action) regarding 9 factors. The order of factors/features was randomized for each participant. Before answering the three questions, participants in the AI Prediction and AI Explanation groups were reminded of the AI system's prediction and in the AI Explanation group people were additionally told the CF explanation for that factor. For example, questions and preceding text relating the education level were as follows:

\begin{itemize}
    \item[] Reminder: The AI system \underline{predicts} that Tom’s yearly salary is \textit{ \textbf{lower than £30k ($<$ £30k)}}. [AI Prediction and AI Explanation groups]
    \item[] Factor: \textbf{Education level} [all three groups]
    \item[] \underline{Explanation}: If Tom had \textbf{had an advanced degree (e.g.~masters)}, the AI system would have predicted that his \textbf{salary} was \textbf{higher than/equal to £30k ($\geq$ £30k)}. [only the AI Explanation group]

\vspace{1em}

    \item[] \textbf{Q.} Would you expect that employees who \textbf{have an advanced degree (e.g.~masters)} also have a \textbf{higher salary}? [Expectation question, same for all three groups]
 
    \item[] Please rate your answer from 0 (No, not at all) to 100 (Yes, absolutely).

    \item[] \textbf{Q.} How \underline{confident} are you in your response? [Confidence question, same for all three groups]

    \item[] \textbf{Q.} Assuming Tom has the resources (time, money, etc.), would you \underline{recommend} he \textbf{starts an advanced degree (e.g.~masters)} with the hope of increasing his \textbf{salary}? [Action question, same for all three groups]

    \item[] Please rate your answer from 0 (not at all) to 100 (totally).

\end{itemize}

\noindent Participants' responses to the three questions were elicited using a slider from 0 to 100 with 1 point increments. The three questions followed the same format for all other factors. The Action questions sometimes had a short caveat (`Assuming Tom has the resources \ldots') as shown above to guard against participants drifting into a cost-benefit analysis which could deter from them providing causal estimates regarding the factor in question. The format of the CF explanations was the same for each factor, namely `If Tom had [had/worked/owned etc.~the factor/feature], the AI system would have predicted that his \textbf{salary} was \textbf{higher than/equal to £30k ($\geq$ £30k)}'. For example, for factor `office plant' the explanation was `If Tom had \textbf{owned an office plant}, the AI system would have predicted that his \textbf{salary} was \textbf{higher than/equal to £30k ($\geq$ £30k)}' and for factor `penthouse apartment' the explanation read `If Tom had \textbf{rented a penthouse apartment}, the AI system would have predicted that his \textbf{salary} was \textbf{higher than/equal to £30k ($\geq$ £30k)}'. Given that this formulation of the CF explanation implies a positive impact of the factor/feature on salary, we expected that participants' Action estimates in the AI Explanation group would be \emph{higher} than the Action estimates of the participants in the other two groups. At the end of the survey participants were asked to summarize their reasoning for the estimates they provided in a free format type text box. This information was used to gain insight into the potential approaches participants took to answer the questions. Lastly, participants received a debriefing information and were invited to provide feedback.

\subsection{Results}

\paragraph{Familiarity with the factors affecting salary and AI systems}

We first analyzed the participants estimates regarding how familiar they are with factors affecting salary. We performed a one-way ANOVA for each familiarity category (i.e.~salary and AI systems) with group (Control, AI Prediction, AI Explanation) as a three-level independent variable. We found no significant effect of group on either familiarity with factors affecting salary, $F(2, 90) = 0.66$, $p = .52$, or familiarity with AI systems, $F(2, 90) = 1.96$, $p = .15$. Mean familiarity ratings indicated that participants were more familiar with factors affecting salary ($M=3.9$) than AI systems ($M=2.8$), which is expected. These results suggests that any potential significant differences between the groups in the further analyses cannot be accounted for by the participants familiarity with the domain (salary) or AI systems.

\paragraph{Main analyses}

\begin{figure*}[h]
    \centering
    \begin{minipage}{0.49\textwidth}
        \centering
        \includegraphics[width=1\textwidth]{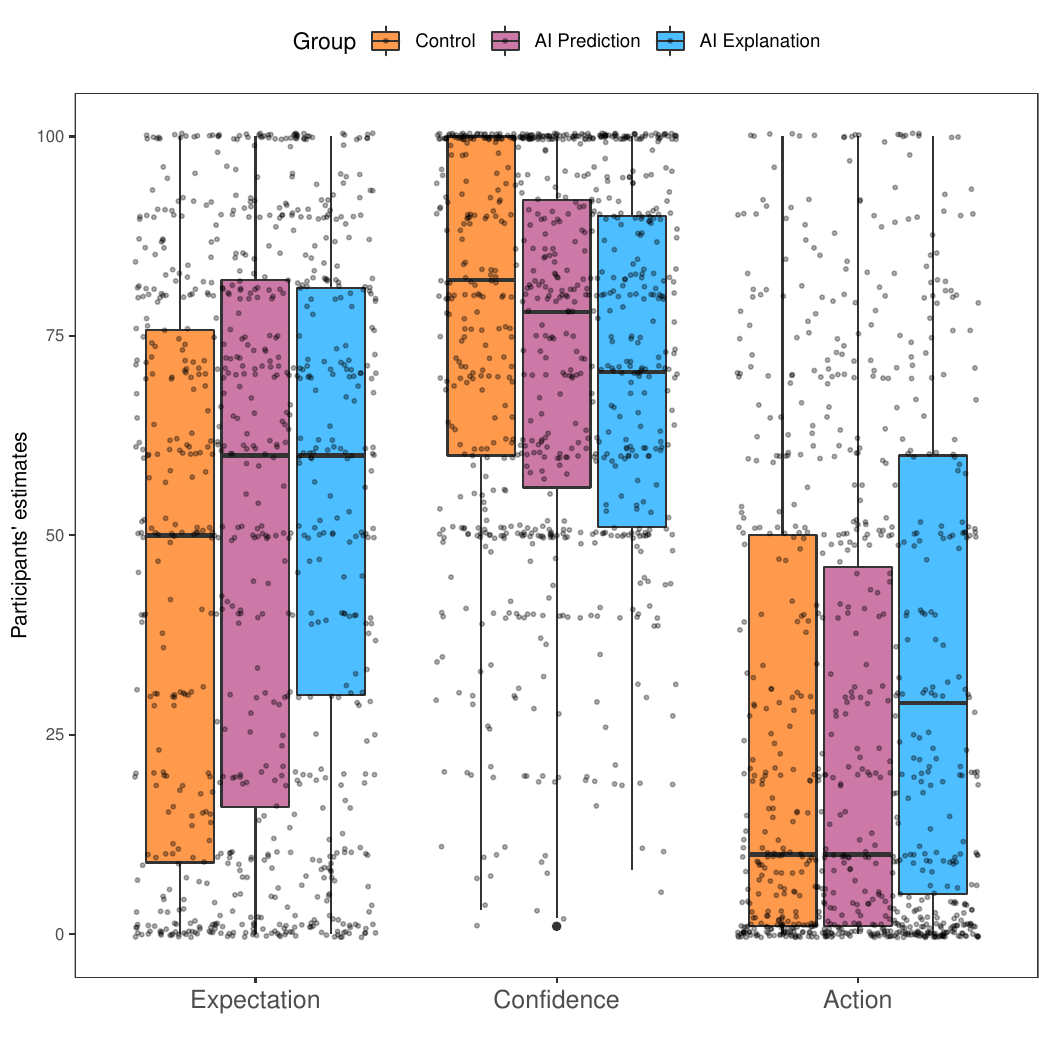} 
        \caption{Experiment 1 results for each each group and each dependent variable. We find that Action estimates were significantly higher in the AI explanation group than in the two other groups. We also find that AI Explanation group's Expectation estimates were not significantly higher than AI Prediction group's estimates, suggesting that the effect of CF explanations on Action estimates is not due to participants beliefs in the AI's predictive power.}
        \label{fig: ex1_box_all}
    \end{minipage}\hfill
    \begin{minipage}{0.49\linewidth}
        \centering
        \includegraphics[width=1\textwidth]{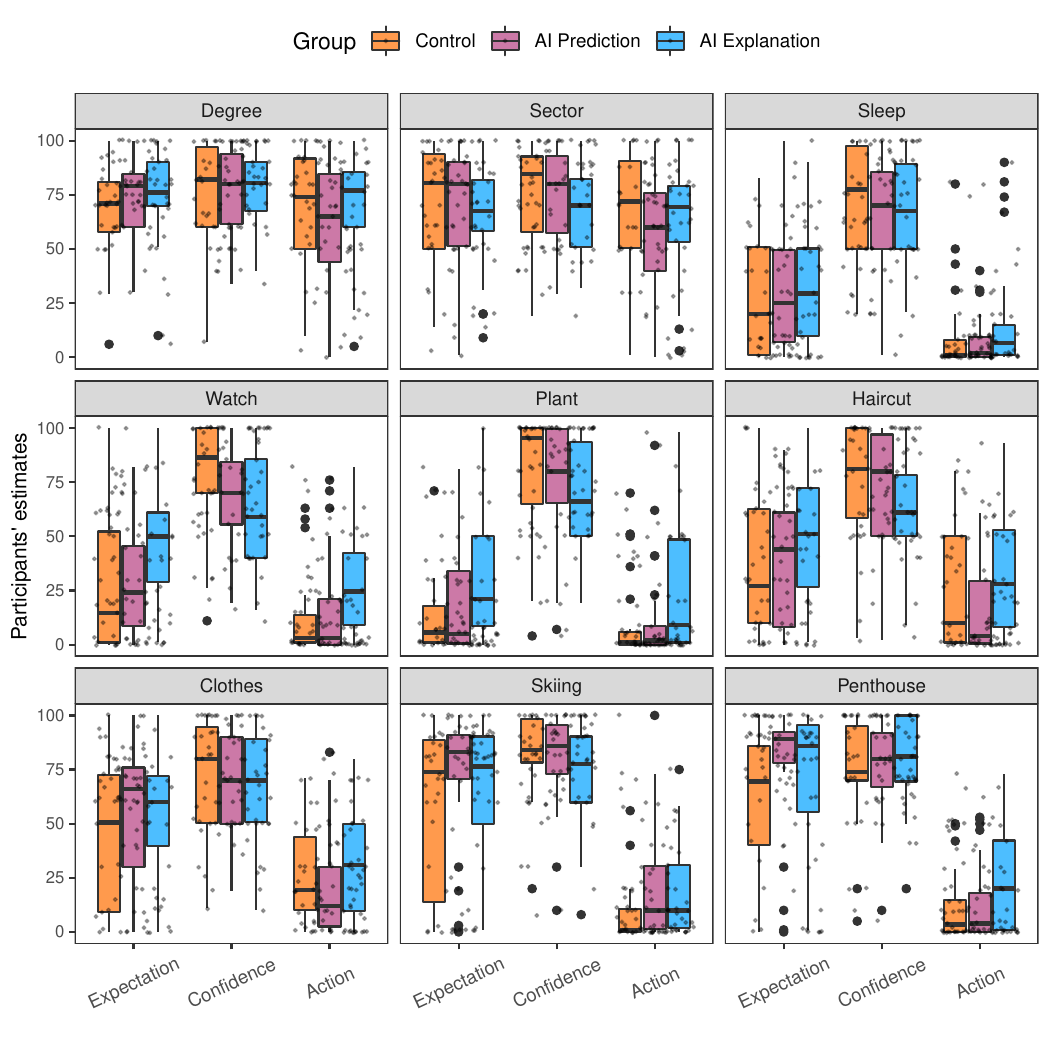} 
        \caption{Experiment 1 results for each factor/feature, each group, and each dependent variable. We find a trend whereby the participants' Action estimates were higher in the AI Explanation group than in the other two groups across the factors. We also find that for factors that are intuitively more causal both Expectation and Action estimates were high, for factors that are intuitively not causal both Expectation and Action estimates were low, and for factors that are intuitively effects rather than causes of higher salary Expectation estimates were high and Action estimates were low, in agreement with the experimental predictions.}
  \label{fig: ex1_box}
    \end{minipage}
\end{figure*}

Participants estimates for each dependent variable are shown in Figure~\ref{fig: ex1_box_all}. To test the effect of group on each dependent variable we initially built three linear mixed-effects models with the random intercept for each participant. However, as the distributions of participants estimates were highly skewed (especially for Expectation and Action dependent variables) and as residuals of the linear mixed-effects models were clearly non-normally distributed (see Appendix B), to test for the overall effect of the group we resorted to non-parametric tests. We performed a Kruskal-Wallis rank sum test for each dependent variable with the group as a three-level independent variable. Participants' expectation estimates were significantly affected by the group they were assigned to (Control, AI Prediction, AI Explanation) for Expectation estimates, $H(2) = 11.9$, $p = .003$, Confidence estimates, $H(2) = 16.1$, $p < .001$, and Action estimates, $H(2) = 27$, $p < .001$.

We performed post-hoc pairwise comparisons between the three groups using Wilcoxon Rank Sum test (the false discovery rate for multiple comparisons was controlled using the Benjamini–Hochberg procedure \cite{benjamini1995controlling}; for more details see Table~\ref{tab:pair_ex1} in Appendix A). We found that participants' Action estimates were not significantly different between Control and AI prediction groups ($p = .74$), but that there was a significant difference between AI Prediction and AI Explanation ($p<.001$) as well as between Control and AI Explanation ($p<.001$). From Figure~\ref{fig: ex1_box_all} we can also see that participants' Action estimates were higher in AI Explanation group compared to the two other groups. Figure~\ref{fig: ex1_box} suggests that this effect held across the features/factors and not just overall. These results provide support for our main hypothesis, namely that providing CF explanations would affect people's beliefs about how causal are the features in the real world.

Post-hoc pairwise comparisons with respect to the participants' Expectation estimates showed significant differences between Control and AI Prediction ($p=.03$) as well as Control and AI Explanation ($p=.002$); however, AI Prediction group's and AI Explanation group's estimates were not significantly different, $p=.36$. These results support our second hypothesis:~being aware that an AI system is using certain factors/feature to make predictions and knowing what the prediction is affects people's expectation as to how well these features/factors are predicting salary. However, their expectations will not further change upon learning about the CF explanations of the features/factors. This also implies that the results regarding participants' Action estimates cannot be explained by the participants' Expectation estimates, providing further support for the claim that CF explanations are affecting people's causal beliefs.

Post-hoc pairwise comparisons on participants' Confidence estimates showed significant difference between Control and both AI Prediction ($p=.02$) and AI Explanation ($p<.001$) groups. There was not a significant difference between AI Prediction and AI Explanation groups ($p=.12$). Figure~\ref{fig: ex1_box_all} shows a downward trend in estimates from the Control to the AI Explanation group. We speculate that this might be because some of the features/factors the AI system uses are intuitively not relevant to salary or they are effects rather than causes of higher/lower salary. This may result in the reduction in people's confidence in the AI system's predictive accuracy. It is important to note that participants' Confidence estimate were clearly different from their Expectation estimates, suggesting that these two dependent variables were successfully disentangled in the experiment design.

Lastly, Figure~\ref{fig: ex1_box} shows that the participants' estimates were dependent on the feature/factor they were asked to provide estimates for. These roughly coincided with the intuitions outlined above, namely that factors/features such as education level and sector would have high both Expectation and Acton estimates as they seem that they can causally affect salary; factors/features such as office plant, and smart watch do not seem causally relevant for salary hence low estimates for both Expectation and Action, and factors/features such as expensive clothes and renting a penthouse apartment are intuitively effects of higher salary, hence their Expectation estimates would be higher, whereas their Action estimates would be generally low. Critically, we found that Action estimates were higher in AI Explanation group compared to other two groups across most of the factors/features. This was not the case for the participants' Action estimates for education level (Degree) and Sector factors. We found that the group participants were assigned to did not significantly affect their Action estimates ($H(2) = 4.2$, $p = .12$.). This is expected for two reasons. For one, as participants in the Control group already have high Action estimates for these two factors (following our conjecture that these two factors are intuitively causal factors), adding a counterfactual explanation may further reinforce but not significantly change their causal beliefs about these two factors. On the other hand, we found that participants' Action estimates were significantly different depending on the group they were assigned to for all other factors ($H(2) = 38.1$, $p < .001$.). Secondly, as our scales was bounded at 100, it is possible that some of the non-significant finding is due to ceiling effects.

\section{Experiment 2}

Experiment 1 suggested that providing lay users with CF explanations of AI systems' predictions can (unjustifiably) affect their causal beliefs about the features/factors. The aim of Experiment 2 experiment was to explore whether we can correct the effects of CF explanations on people's causal beliefs. Inspired by the research on correcting misinformation \cite{irving2022correcting} and the research on the impact of health warning messages \cite{hammond2011health}, we designed this experiment to explore whether providing participants with a note communicating that AI systems are capturing correlations in data rather than causal relationships might attenuate the effect of CFs on their causal beliefs. We hypothesise that the AI Explanation group presented with the note will provide lower Action estimates than the AI Explanation group where the note was not present. We do not have a specific hypothesis as to how introducing the note might affect participants Expectation, Confidence, or Action estimates in the other groups or how AI Explanation groups' Expectation and Confidence estimates might change due to the note. 

The second aim of Experiment 2 is to provide a replication of Experiment 1 in groups that are not presented with the note. Thus Experiment 2 will provide additional test for the two hypotheses explored in Experiment 1.

\subsection{Methods}
Effect size calculations showed that the effect size of Experiment 1 results was relatively small ($\eta^2=.03$), making Experiment 1 being underpowered. To increase the power of Experiment 2 we increased the number of participants. We aimed to have around 45 participants in each group.

\paragraph{Participants, Design, \& Materials}

A total of 271 participants ($N_{\mathrm{female}} = 196$, two participants identified as neither male nor female, $M_{\mathrm{age}} = 38.7$, $SD=12.2$) were recruited from Prolific Academic (\url{www.prolific.ac}). All participants were native English speakers residing in the UK or Ireland whose approval ratings were 95\% or higher. They all gave informed consent and were paid £6.24 an hour rate for partaking in the present study, which took on average 8.6 min to complete. Participants were randomly assigned to one of 3 (Control, AI Prediction or AI Explanation) $\times$ 2 (correction:~No Note or Note) = 6 between-participant groups (Control \& No Note $N=46$, Control \& No Note $N=45$, AI Prediction \& No Note $N=44$, AI Prediction \& Note $N=46$, AI Explanation \& No Note $N=46$, AI Explanation \& No Note $N=44$). Experiment 2 used the same three dependent variables as Experiment 1, namely Expectation, Confidence, and Action. Materials in Experiment 2 were exactly the same as those in Experiment 1.

\paragraph{Procedure} Experiment 2 procedure was similar to Experiment 1 procedure. The only difference is that three of the 6 groups were additionally presented with a note regarding correlation, causation, and AI systems.  For groups with the note, that note was introduced in the preamble of each condition, presented on a separate page and participants were also reminded of the note before answering the questions related to the three dependent variables. The note read slightly differently for Control, AI Prediction or AI Explanation groups. Control group:

\begin{itemize}
\item[] \textbf{\underline{Important note}}
 
\item[]\textit{Correlation does not imply causation}. Even though some factors may be highly \textbf{correlated} with higher salary that \textit{\textbf{does not}} mean that they are \textbf{causing} higher salary.
\end{itemize}

\noindent The note does not mention the AI system as participants in this group were not presented with any AI system. Instead, the note included general information about correlation and causation. In the AI Prediction group the note read:

\begin{itemize}
\item[] \textbf{\underline{Important note}}
 
\item[] AI systems learn \textit{\textbf{correlations}} in data. Even though the factors the \underline{AI system} uses are potentially \textit{\textbf{correlated}} with higher salary that \textbf{does not} mean that they are \textit{\textbf{causing}} higher salary.
\end{itemize}

\noindent Here participants are told information regarding correlation and causation that is more relevant to the AI systems. Specifically, they are told that AI systems capture relationships that are correlational and should not be interpreted as causal. In the AI Explanation condition the note read:

\vspace{2em}
\begin{itemize}
\item[] \textbf{\underline{Important note}}

\item[] AI systems learn \textit{\textbf{correlations}} in data. Even though the factors the \underline{AI system} uses are potentially \textit{\textbf{correlated}} with higher salary that \textbf{does not} mean that they are \textit{\textbf{causing}} higher salary. Similarly, the \textbf{explanations} of the AI systems' predictions are about the \textit{correlations} the \underline{AI system} has identified and not about which factors are \textit{actually causing} higher salary.
\end{itemize}

\noindent In addition to being told that AI systems capture correlations, participants in this group were also told that the explanations of the AI system’s predictions are explanations of these correlations and not necessarily of causal relations.

\subsection{Results}

\paragraph{Familiarity with the factors affecting salary and AI systems}

Like in Experiment we first analyzed the participants estimates regarding how familiar they are with factors affecting salary.
We performed a two-way ANOVA for each familiarity category (i.e.~salary and AI systems) with group and correction as two factors. We found no significant effect of group (Control, AI Prediction, AI Explanation) on either familiarity with factors affecting salary, $F(2, 265) = 0.99$, $p = .37$, or familiarity with AI systems, $F(2, 265) = 0.13$, $p = .88$. We found no significant effect of correction (No Note, Note) on either familiarity with factors affecting salary, $F(1, 265) = 0.55$, $p = .46$, or familiarity with AI systems, $F(1, 265) = 0.06$, $p = .8$. Finally, we found no significant interaction effect between the two independent variables on either familiarity with factors affecting salary, $F(2, 265) = 1.68$, $p = .19$, or familiarity with AI systems, $F(2, 265) = 1.19$, $p = .31$. Mean familiarity ratings indicated that participants were more familiar with factors affecting salary ($M=3.9$) than AI systems ($M=2.9$). These results are very similar to those in Experiment 1.

\paragraph{Main analyses}

\begin{figure*}[h]
    \centering
    \begin{minipage}{0.495\textwidth}
        \centering
        \includegraphics[width=1\textwidth]{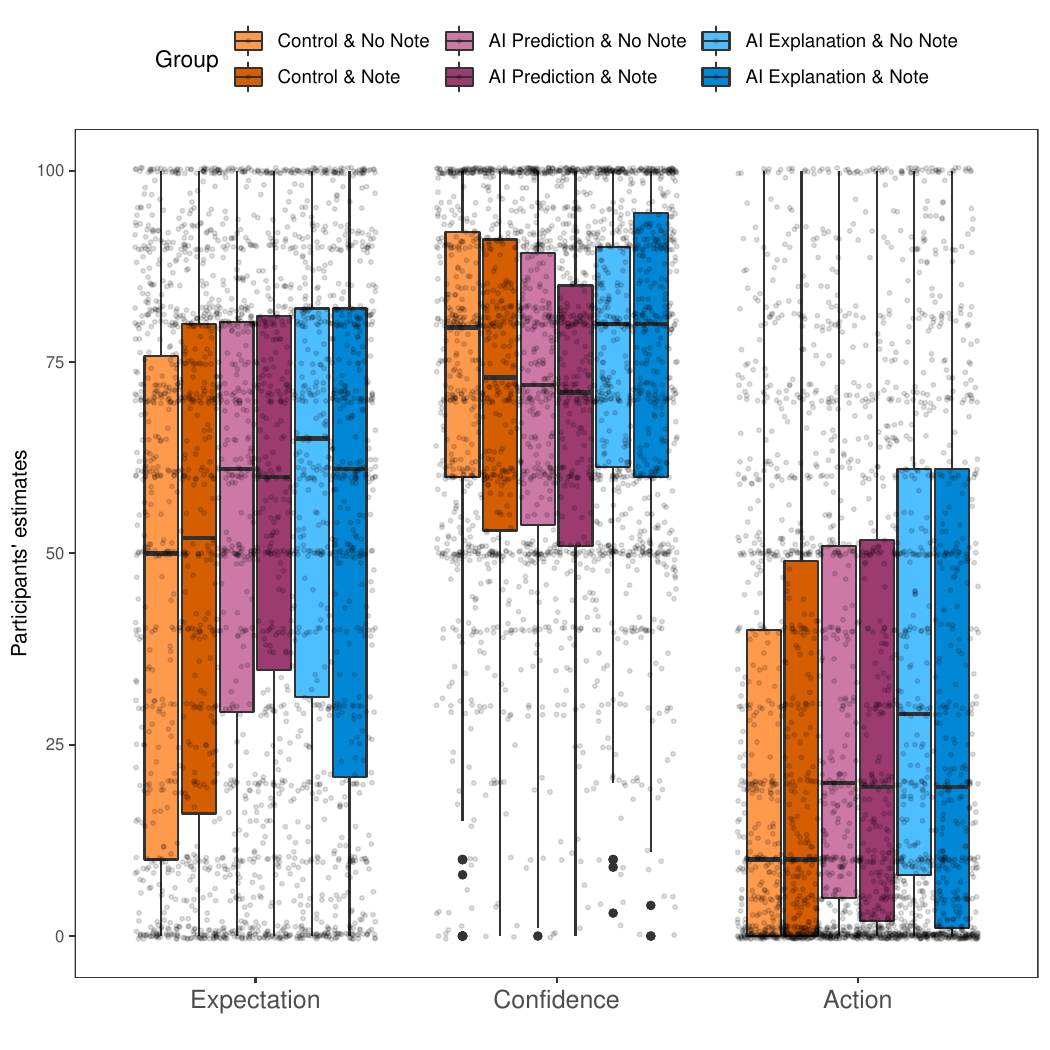} 
        \caption{Experiment 2 results for each dependent variable. Like in Experiment 1 we find that Action estimates are higher in the AI Explanation group and cannot be explained by Expectation estimates when the note is not communicated to the participants. When the note is presented to the participants, Action estimates in AI Explanation group are at the level of AI Prediction group and not significantly higher.}
        \label{fig: ex2_box_all}
    \end{minipage}\hfill
    \begin{minipage}{0.495\linewidth}
        \centering
        \includegraphics[width=1\textwidth]{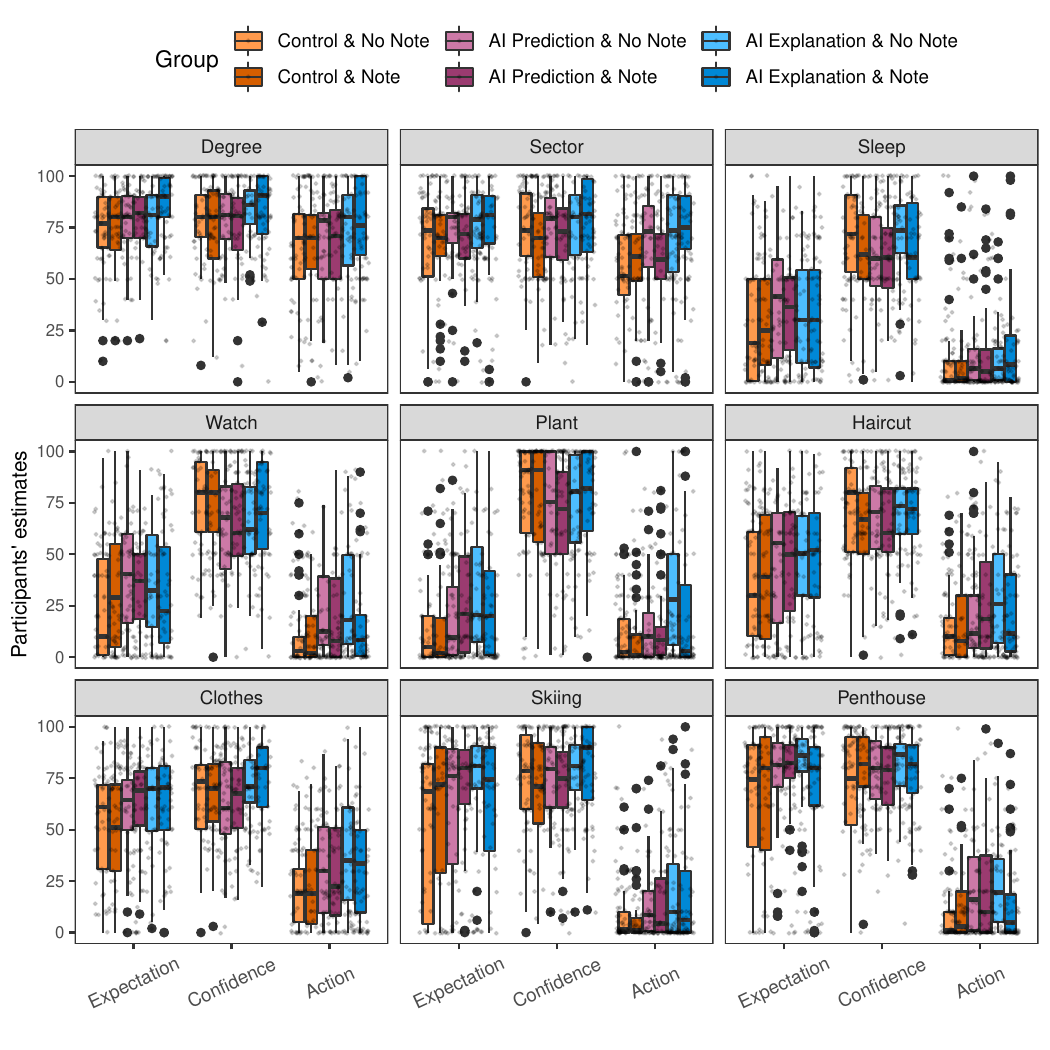} 
        \caption{Experiment 2 results for each factor/feature and each dependent variable. We find that the similar distributions of estimates for different factors as in Experiment 1. Furthermore, we also find the trend of reduction in Action estimates when the note is present compared to when it is not.}
        \label{fig: ex2_box}
    \end{minipage}
\end{figure*}

Participants estimates for each dependent variable as shown in Figure~\ref{fig: ex2_box_all}. Similarly to Experiment 1, the distributions of participants estimates were skewed (especially for Expectation and Action dependent variables) and residuals of the linear mixed-effects models were clearly non-normally distributed (see Appendix B). So, to test for the overall effect of group we performed a Kruskal-Wallis rank sum test for each dependent variable. Participants' expectation estimates were significantly affected by group (Control \& No Note, Control \& Note, AI Prediction \& No Note, AI Prediction \& Note, AI Explanation \& No Note, AI Explanation \& Note) for Expectation estimates, $H(5) = 38.9$, $p < .001$, Confidence estimates, $H(5) = 33.8$, $p < .001$, and Action estimates, $H(5) = 67.7$, $p < .001$.

From Figure~\ref{fig: ex2_box_all} we can also see that participants' Action estimates were significantly higher in AI Explanation \& No Note group compared to the two other No Note groups. This mirrors the findings from Experiment 1 and further supports the first hypothesis from Experiment 1. Unlike in Experiment 1, the difference between Control and AI Prediction condition was also significant, $p=.02$ (see for more details see Table \ref{tab:pair_ex2_act} in Appendix A). Pairwise comparisons for depended variable Expectation show significant difference only between Control and both AI Prediction ($p = .001$) and AI Explanation ($p<.001$) groups. No significant difference was found between AI Prediction and AI Explanation ($p=.31$). This result provides support to our second hypothesis from Experiment 1 and suggests that even though there were significant difference between all three No Note groups in Action dependent variable, the significant difference between AI Prediction and AI Explanation cannot be accounted for by differences in Expectation estimates.

Pairwise comparisons across all three dependent variables show that the only significant difference between No Note and Note conditions was between AI Explanation \& No Note and AI Explanation \& Note for Action dependent variable, $p=.01$ (see Tables \ref{tab:pair_ex2_exp}, \ref{tab:pair_ex2_conf}, and \ref{tab:pair_ex2_act} in Appendix A for more details). Participants' Action estimates in group AI Explanation \& Note were lower than those in group AI Explanation \& No Note and were not significantly different than those from AI Prediction \& No Note or AI Prediction \& Note. This implies that the effect of CF explanations on participants causal belief was attenuated and no different from that in groups where CF explanations of AI systems' predictions were not shown. Moreover, Figure \ref{fig: ex2_box_all} suggests that participants' Action estimates in AI Explanation \& Note were lower than those in Explanation \& No Note for almost all features/factors. These results directly support our hypothesis.

Finally, participants' Confidence estimates were again different from their Expectation estimates. But, unlike in Experiment 1 where there was a downward trend in participants' Confidence estimates, Experiment 2 found that AI Prediction groups' estimates were lower than both Control groups' and AI Explanation groups' estimates, and that there was no significant difference between Control groups' and AI Explanation groups' estimates. In Experiment 1 we speculated that  Confidence estimates might be driven by some factors/features not being relevant to salary or in an anti-causal relationship to (i.e.~effects of) salary. However, the data form Experiment 2 does not seem to support this supposition.

\section{Discussion}

If one of the aims of explainable AI is to provide human users with information that will help them better understand how an AI system came to a prediction and how the system will behave in the future, then we need to communicate to that user as clearly as possible the predictive and associative (rather than the causal) nature of these systems so that the mental models humans create based on that information are more genuine and representative of the AI system's nature.

Two experiments showed that participants' causal estimates were significantly higher when they were presented with a CF explanation compared to both the baseline and when only the prediction was communicated. We further found that this was not the case for people's beliefs regarding how good are the feature/factors in predicting salary and that there was no significant difference in expectation estimates difference between the group where only predictions was presented and the group where both the prediction and a CF explanation was included. This result suggests people's expectation estimates cannot account for the differences in their causal beliefs and that these differences were in fact due to CF explanations alone. This implies that CF explanation of AI systems' predictions can (unjustifiably) skew people's causal beliefs about the world.

We also found that one might be able to guard against the unwanted effect of CF explanations on causal beliefs. Inspired by the work on misinformation and health warning messaging, we designed a note communicating to the participants the correlational character of AI systems rather than causal. Adding the note reduced the effect of CF explanation on the participants' causal beliefs.

\subsection{Future work}\label{s: limit}

In this study we have used salary as a domain. This is because we expected that participants are largely familiar with this domain, which would allow us to test whether their Expectation and Action estimates were in line with what we expected. As their estimates were in line with our expectations, this provided evidence for the appropriateness of the metrics we used. Further research should explore other domains; in particular, the domains that people are not as familiar with. We expect, however, that the effect of adding a CF explanation on Action estimates will be at least as great as in this study. From Figures \ref{fig: ex1_box} and \ref{fig: ex2_box} we can see that when participants' Expectation estimates were relatively low (e.g.~owning a smart watch, owning an office plant, and getting expensive haircuts), implying that they believed the association between the features/factors and salary was weak, their Action estimates were invariably higher in AI Explanation condition compared to the other two conditions. If, as it seems plausible, when reasoning in a relatively unfamiliar domain participants believe there is a weak association between the features/factors and the label, then we would expect to find a greater impact of adding CF explanations on their Action estimates compared to this study as this study also included features that are intuitively causally efficacious with respect to the label (e.g.~having an advance degree or sector the employee works in).

A wealth of research on explanation and explanatory goodness suggests that simpler explanations have a bigger impact on our (causal) beliefs \cite{lombrozo2007simplicity,lagnado1994,read1993explanatory,thagard1978best}. In our studies only one feature/factor was included in a CF explanation at a time, so our CF explanations were on the simpler side of the spectrum. This could imply that the CF explanations that include multiple factors and are a combination of these factors have less impact on our causal beliefs about the world than the simpler CF explanation that use one or two factors. Further research should explore how more complex CF explanations of AI systems' predictions affect people's causal beliefs about the world. It should be noted that although increasing the complexity of CF explanations may reduce their undesired impact on our causal beliefs, it may at the same time increase the time needed to process these explanations and reduce satisfaction \cite{narayanan2018humans}.

Finally, we have only briefly discussed the role of the participants' confidence in their expectation estimates. We found that Confidence estimates are clearly different from the expectation ones. However, we have not explored in further detail how confidence estimates may depend on whether people are just told about the AI system's prediction or they are also told the CF explanation. It may be interesting to explore how  confidence estimates interact with people's estimates of how accurate they believe the AI system is in predicting the label.


%

\bibliographystyle{named}
\bibliography{refs}

\appendix

\section{Pairwise comparisons}

Table \ref{tab:pair_ex1} shows post-hoc pairwise comparison for each dependent variable in Experiment 1. Tables \ref{tab:pair_ex2_exp}, \ref{tab:pair_ex2_conf}, and \ref{tab:pair_ex2_act} include all post-hoc pairwise comparisons for each dependent variable in Experiment 2.

\begin{table*}
  \caption{Experiment 1:~Pairwise Wilcoxon Rank Sum tests $p$-values for all three dependent variables. All $p$-values were corrected for multiple comparisons using Benjamini and Hochberg's false discovery rate (FDR) procedure \protect\cite{benjamini1995controlling}.}
  \label{tab:pair_ex1}
  \begin{tabular}{lcccccc}
    \toprule
    &\multicolumn{2}{c}{Expectation}&\multicolumn{2}{c}{Confidence}&\multicolumn{2}{c}{Action}\\
    &Control&AI Prediction&Control&AI Prediction&Control&AI Prediction\\
    \midrule
    AI Prediction & .03 & & .02 & & .74 \\
    AI Explanation & .002 & .36 & < .001 & .12 & < .001 & < .001  \\
  \bottomrule
\end{tabular}
\end{table*}

\begin{table*}
  \caption{Experiment 2:~Pairwise Wilcoxon Rank Sum tests $p$-values for dependent variable Expectation. All $p$-values were corrected for multiple comparisons using the false discovery rate method.}
  \label{tab:pair_ex2_exp}
  \begin{tabular}{lccccc}
    \toprule
    &Control&Control&AI Prediction&AI Prediction&AI Explanation\\
    &\& No Note&\& Note&\& No Note&\& Note&\& No Note\\
    \midrule
    Control \& Note & .33& \\
    AI Prediction \& No Note & .001 & .02 \\
    AI Prediction \& Note & < .001 & .004 & .68  \\
    AI Explanation \& No Note & < .001 & < .001 & .31 & .47  \\
    AI Explanation \& Note  & < .001 & .01 & .65 & .88 & .59\\
  \bottomrule
\end{tabular}
\end{table*}

\begin{table*}
  \caption{Experiment 2:~Pairwise Wilcoxon Rank Sum tests $p$-values for dependent variable Confidence. All $p$-values were corrected for multiple comparisons using the false discovery rate method.}
  \label{tab:pair_ex2_conf}
  \begin{tabular}{lccccc}
    \toprule
    &Control&Control&AI Prediction&AI Prediction&AI Explanation\\
    &\& No Note&\& Note&\& No Note&\& Note&\& No Note\\
    \midrule
    Control \& Note & .08& \\
    AI Prediction \& No Note & .03 & .77 \\
    AI Prediction \& Note & .001 & .17 & .28  \\
    AI Explanation \& No Note & .95 & .08 & .03 & < .001  \\
    AI Explanation \& Note  & .31 & .008 & .003 & < .001 & .28\\
  \bottomrule
\end{tabular}
\end{table*}

\begin{table*}
  \caption{Experiment 2:~Pairwise Wilcoxon Rank Sum tests $p$-values for dependent variable Action. All $p$-values were corrected for multiple comparisons using the false discovery rate method.}
  \label{tab:pair_ex2_act}
  \begin{tabular}{lccccc}
    \toprule
    &Control&Control&AI Prediction&AI Prediction&AI Explanation\\
    &\& No Note&\& Note&\& No Note&\& Note&\& No Note\\
    \midrule
    Control \& Note & .63& \\
    AI Prediction \& No Note & < .001 & < .001 \\
    AI Prediction \& Note & .001 & .003 & .39  \\
    AI Explanation \& No Note & < .001 & < .001 & .02 & .001  \\
    AI Explanation \& Note  & < .001 & .002 & .63 & .63 & .01\\
  \bottomrule
\end{tabular}
\end{table*}

\section{Linear mixed effects models}

\subsection{Experiment 1}

To estimate the effect of group on the three dependent variables we initially built linear mixed-effects models (LMM) using the ``lme4'' package in R \cite{bates2014}. The only fixed effect was group (with three levels:~Control, AI Prediction, AI Explanation). The only random effect was the intercept for participants. There was no random slope from the participant as the design was fully between. No random intercept for scenarios was used as the number of scenarios was low (i.e.~9) and including the scenarios as a random intercept could have lead to a reduced power of the experiment \cite[see]{Judd2017,singmann2019}. Further, a random slope for scenarios was not included as led to a singular fit model, implying that the variance of this random effect was (close to) zero.

\begin{figure*}
    \centering
    \begin{minipage}{0.49\textwidth}
        \centering
        \includegraphics[width=\linewidth]{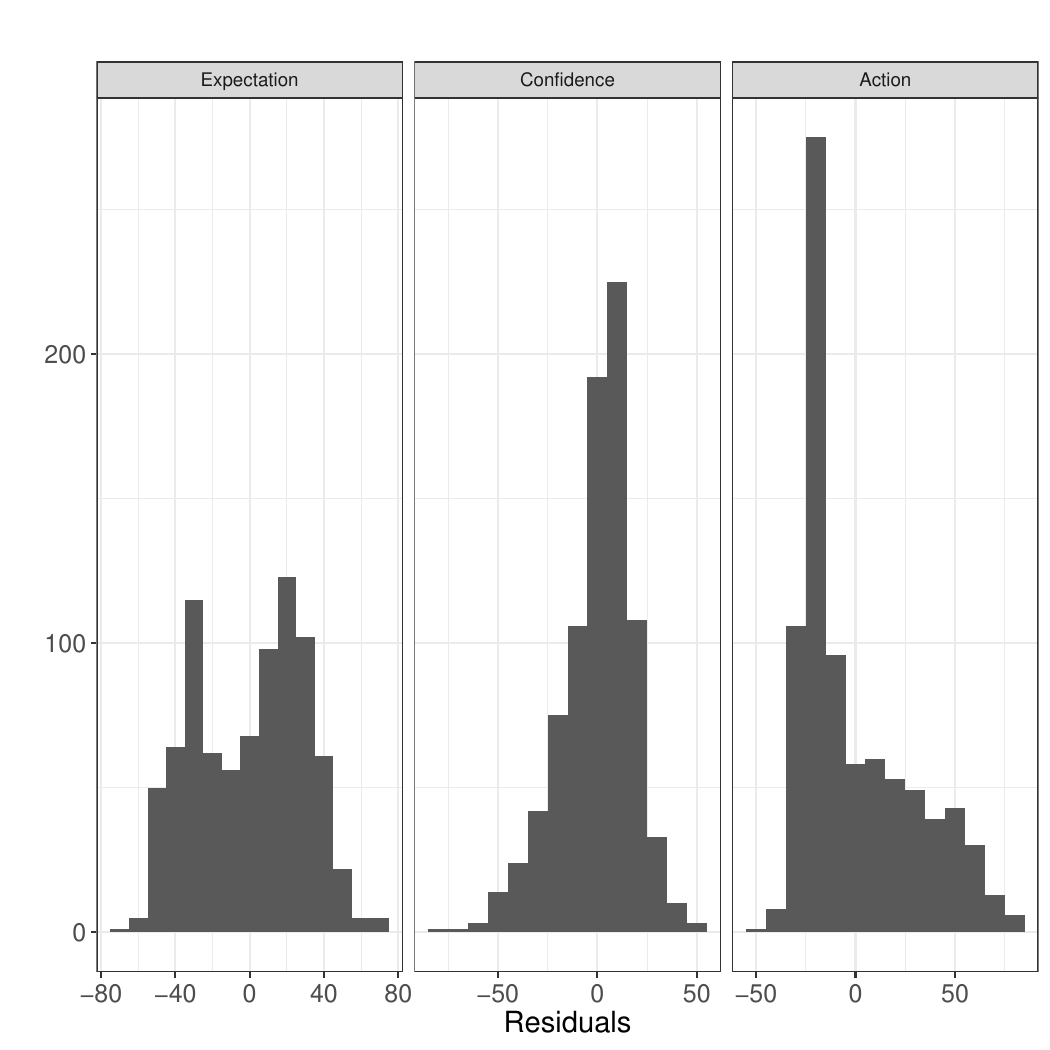}
  \caption{Histograms of the LMM residuals for all three dependent variables in Experiment 1.}
  \label{fig: ex1_hist_res}
    \end{minipage}\hfill
    \begin{minipage}{0.49\linewidth}
        \centering
        \includegraphics[width=\linewidth]{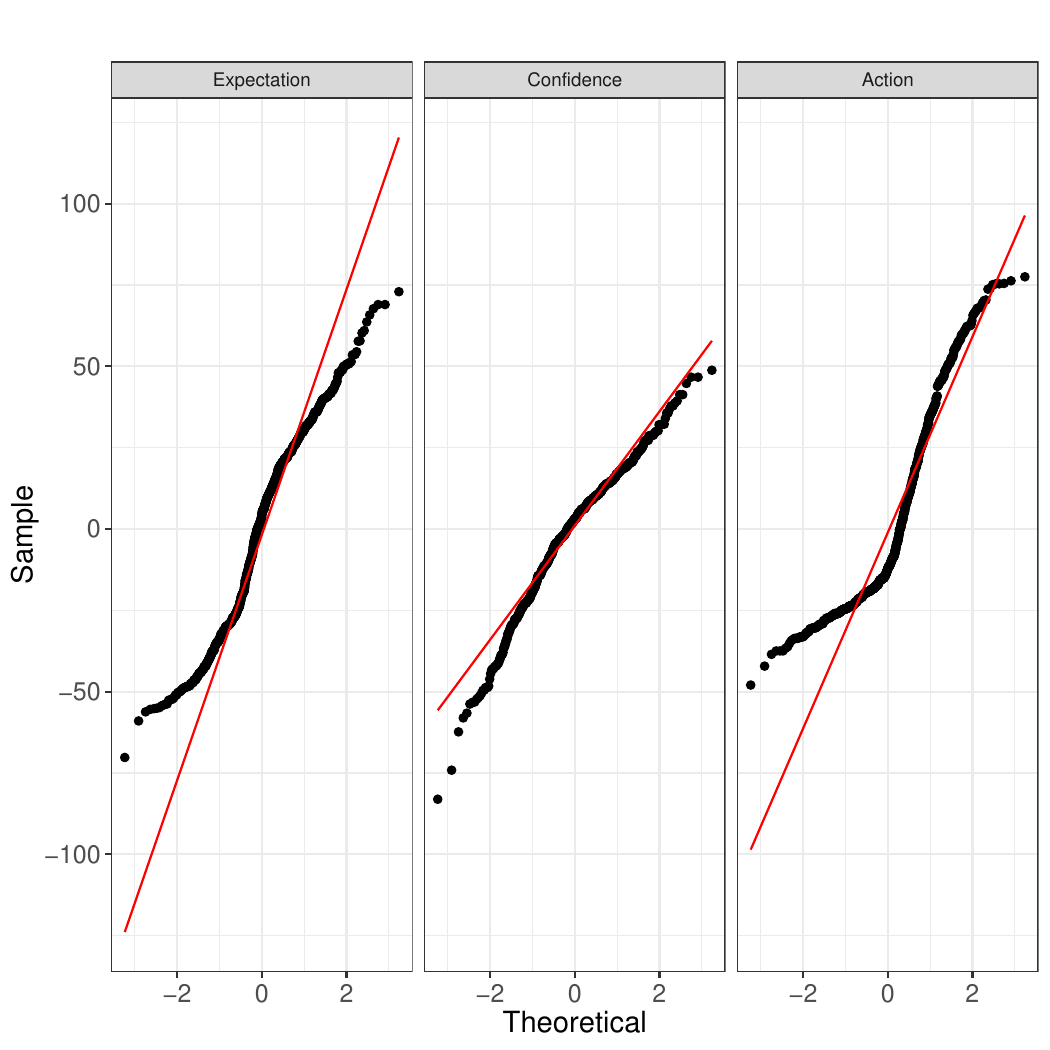}
  \caption{Quantile plots of the LMM residuals for all three dependent variables in Experiment 1.}
  \label{fig: ex1_quant_res}
    \end{minipage}
\end{figure*}

After we fitted the LMM, we plotted the quantile plots of the residuals and the histograms of residuals. Figures \ref{fig: ex1_hist_res} and \ref{fig: ex1_quant_res} show that the residuals are non-normally distributed for all dependent variables. Consequently, we resorted to the non-parametric statistical analyses outlined in the main text.

\subsection{Experiment 2}

\begin{figure*}
    \centering
    \begin{minipage}{0.49\textwidth}
        \centering
        \includegraphics[width=\linewidth]{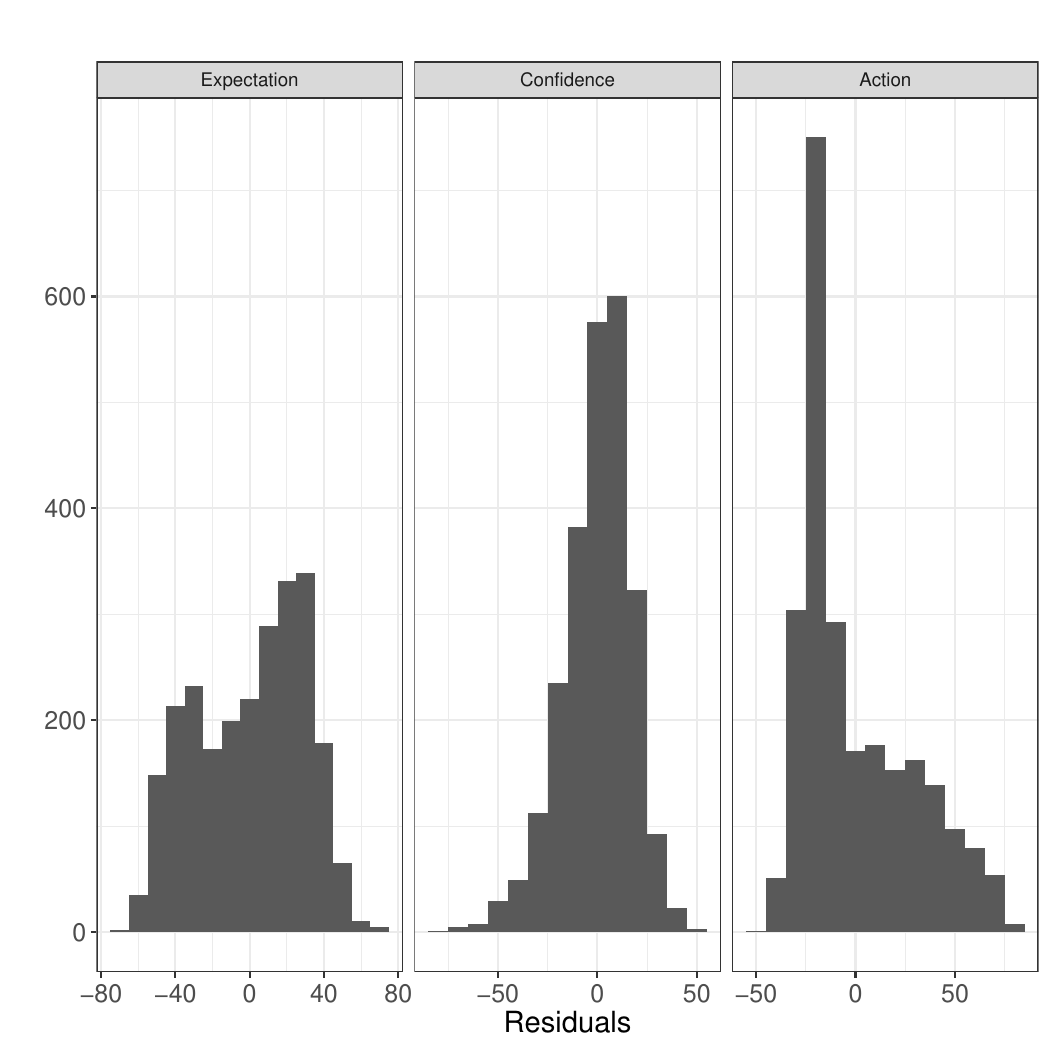}
  \caption{Histograms of the LMM residuals for all three dependent variables in Experiment 2.}
  \label{fig: ex2_hist_res}
    \end{minipage}\hfill
    \begin{minipage}{0.49\linewidth}
        \centering
        \includegraphics[width=\linewidth]{Exp_1_quant_plot_resid.pdf}
  \caption{Quantile plots of the LMM residuals for all three dependent variables in Experiment 2.}
  \label{fig: ex2_quant_res}
    \end{minipage}
\end{figure*}

We build a similar LMM for Experiment 2. The only difference was that instead of only one fixed effect we now had two:~condition (Control, AI Prediction, AI Explanation) and correction (No Note, Note). The random effects structure was the same as in Experiment 1. We again plotted the residuals and found that they were not normally distributed (see Figures \ref{fig: ex2_hist_res} and \ref{fig: ex2_quant_res}). We then performed the same non-parametric analyses as in Experiment 1.

\end{document}